\documentclass[journal,10pt,final]{IEEEtran}
\usepackage{color}
\usepackage{tabulary}
\usepackage{tabu}
\usepackage{multirow}
\usepackage{booktabs}
\usepackage{subcaption}

\usepackage{ifpdf}

\usepackage{cite}

\ifCLASSINFOpdf
  \usepackage[pdftex]{graphicx}
\else
  \usepackage[dvips]{graphicx}
\fi
\usepackage{epsfig}

\usepackage[cmex10]{amsmath}
\usepackage{amssymb}
\begin{document}
%
\title{Transfer Learning for Video Recognition\\with Scarce Training Data\\for Deep Convolutional Neural Network}
%
%
%

\author{
Yu-Chuan~Su,
Tzu-Hsuan~Chiu,
Chun-Yen~Yeh,
Hsin-Fu~Huang,
Winston~H.~Hsu
}

\maketitle

\begin{abstract}
Recognizing high level semantic concepts and complex events in consumer videos has become an important research topic 
with great application needs in the past few years.
In this work, we apply Deep Convolution Network (DCN), the state-of-the-art static image recognition algorithm,
to recognize semantic concepts in unconstrained consumer videos.
Our preliminary studies show that the annotation of video corpora is not sufficient to learn robust DCN models.
The networks trained directly on the video dataset suffer from significant overfitting.
The same lack-of-training-sample problem limits the usage of deep models on a wide range of computer vision problems 
where obtaining training data are difficult.
To overcome the problem, we perform transfer learning from weakly labeled images corpus to videos.
The image corpus helps to learn important visual patterns for natural images, 
while these patterns are ignored by the models trained only on the video corpus.
Therefore, the resultant networks have better generalizability and better recognition rate.
By transfer learning from image to video, we can learn a video recognizer with only 4k videos.
Because the image corpus is weakly labeled, the entire learning process requires only 4k annotated instances,
which is far less than the million scale image and video datasets used in previous works.
The same approach can be applied to other visual recognition tasks where only scarce training data is available.
Our extensive experiments on network configurations also lead to better understanding about the effect of various meta-parameters in DCN,
which enable better network architecture design for future researches and applications.
\end{abstract}


%
\IEEEpeerreviewmaketitle

\section{Introduction}

Semantic concept and complex event recognition in unconstrained consumer videos has received much research interest in recent years 
\cite{jiang:2011icmr,izadinia2012eccv,xu2014arxiv,jiang2015arxiv,wu2015arxiv,ye2015arxiv}
due to both the applicability of technology and real application needs.
Thanks to the popularity of video capturing devices such as smart phones and multimedia sharing sites such as Youtube, 
people nowadays generate and share a large volume of videos over the internet everyday.
It becomes an urgent need to automatically organize and manage these videos, 
and video recognition is a promising first step for intelligent video management systems.
Many previous works focus on recognition in very specific domains such as action recognition \cite{Chaquet:2013},
and the videos usually come from limited sources such as news, movies or recorded in lab environment \cite{Smeaton:2006,Laptev:2008}.
Recognizing complex events in consumer video is a more challenging task, 
because it requires identifying various semantic concepts that may appear in various spatio-temporal locations in the video,
in contrast to the temporally local action recognition.
Also, the videos are captured in more diverse environments, 
which require more discriminative representations for recognition.

\begin{figure}[!t]
\centering
\includegraphics[width=\linewidth]{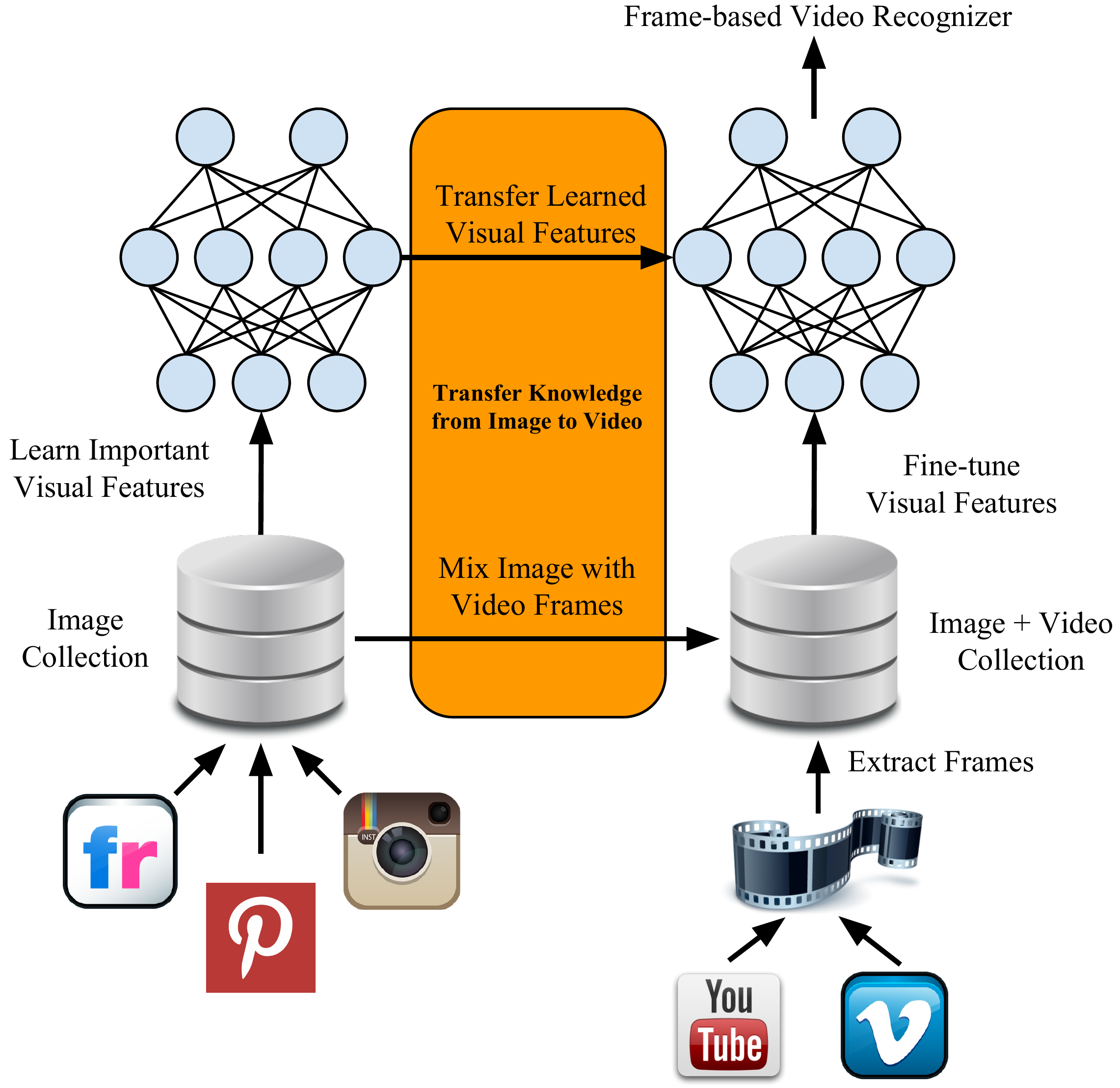}
\caption{
\label{fig:system_diagram} The work flow of learning Deep Convolution Network (DCN) for unconstrained video recognition using transfer learning.
When applying DCN to video recognition, it suffers from the lack-of-training-sample problem which leads to significant overfitting,
while obtaining a large volume of videos with label information is non-trivial.
To overcome the problem, we incorporate the knowledge from weakly labeled image collections through transfer learning from image to video.
The image corpus helps DCN to learn more generalizable features
by transferring the visual appearance learned on the image corpus and augment the training data,
while it requires nearly zero human knowledge and intervention.
It makes learning DCNs with scarce training data possible.
}
\end{figure}

Extensive efforts have been made to design video features for video recognition.
One common approach is to utilize image-based features to capture appearance information in each frame
and then
aggregate multiple frames with video-specific pooling method \cite{jiang:2011icmr,pirsiavash2012cvpr,xu2014arxiv}.
In addition to image-based features,
effort has been made to develop video specific features that capture information encoded in the temporal dimension \cite{Scovanner:2007,klaser:2008}.
These features have improved video recognition performance in certain domains,
but they also introduce significant computation overhead.
Moreover, the performance gain of using video specific features is task and dataset dependent.
In fact, some empirical results show that image-based features may perform as well as or even more robust than spatio-temporal features,
especially when detecting complex events or high level semantic concepts in the video \cite{jiang:2011icmr,pirsiavash2012cvpr,Karpathy:2014cvpr,zha2015arxiv}.
Therefore, designing discriminative video features remains the key challenge for video recognition.

Motivated by recent successes of DCN in concept recognition and object detection on static image,
we aim to exploit DCN to extract better video representation.
Deep Learning as a learning paradigm has been shown effective in various domains including natural language processing \cite{Deselaers:2009}, 
speech recognition \cite{Mohamed:2012}, etc.
In computer vision, 
the great success of DCNs on the ImageNet dataset \cite{Krizhevsky:2012nips,Le:2012icml} has attracted extensive research interest,
and many further efforts have made DCN the state-of-the-art algorithm for various visual recognition tasks in static images.
However, there remains several difficulties for applying DCNs on visual recognition in practice.

The first difficulty is its dire need for extremely large amount of labeled training data.
For example, the ImageNet dataset that is widely used as the test bed as well as the source for pre-training DCNs 
contains more than a million images, all labeled by human.
Collecting such a large dataset with complete ground truth is very challenging and may not be possible in some domains.
This is especially true for video recognition,
where most existing video corpora are one to two order smaller than those used for static image recognition.
Although efforts have been made to collect million scale video recognition benchmarks \cite{Karpathy:2014cvpr},
the label of the dataset is still in video level.
Because irrelevant contents usually interleave with important information in user generated videos,
the video level annotation can be considered as a noisy label for visual content in the video and impose additional difficulty for learning DCNs.
Frame level or even pixel level annotation is more precise and informative, but it requires extensive human labor and is unlikely to be scalable.

The second difficulty is the large number of meta-parameters in DCN.
While models such as SVM and logistic regression usually have only two to three meta-parameters in practice
and can be optimized easily using cross validation grid search, the number of meta-parameters in DCN is at least an order larger.
Combined with the extensive computation requirement, grid search for meta-parameters is generally infeasible,
and many previous works simply apply the setting of existing models which yield sub-optimal performance.

In this work, we focus on solving the two problems to enable practical video recognition algorithm using DCN.
To overcome the lack-of-training-data problem and avoid overfitting, 
we apply two transfer learning approaches to transfer knowledge from static image to video.
The transfer learning approach learns important visual patterns from static images and improves the generalizability of DCNs on unseen videos.
To determine the network configuration, 
we perform extensive experiments on independent image datasets.
These experiments lead to better understanding about the correlation between meta-parameters and DCN performance,
which can serve as the basis for heuristic search in both our experiments and any future research in DCN.

The workflow of our video recognition algorithm is shown in Fig.~\ref{fig:system_diagram}.
Motivated by the competitive performance of static image feature in previous research \cite{jiang:2011icmr},
we apply DCN on frame basis and aggregate frame-wise result using late fusion.
We use the image dataset to initialize the network with meaningful visual pattern in static image and augment the video dataset.
These transfer learning approaches lead to better DCN models and improve recognizer performance.
While some recent works exploit various aggregation methods such as recurrent neural network (RNN) \cite{xu2014arxiv,wu2015arxiv} over DCN,
their models are still based on frame-wise DCN and can benefit from better DCN models obtained by our transfer learning approach.

We verify the efficacy of the transfer learning approach on two video datasets that are popular for high level semantic concept recognition.
We also show that important knowledge can be extracted from image corpora without any human labeling effort.
While fully human annotated image corpora lead to better performance,
even image datasets containing only weak supervision collected from the Internet can significantly boost the video recognition rate.
In practice, we can learn robust video recognizers using unlabeled image dataset and 4k annotated videos labeled in frame-level.
The annotation effort is much less than existing works on image recognition.
Our main contribution is that we successfully learn a DCN with reasonable performance using only scarce training data.
With the transfer learning approach, we can apply DCN in more general visual recognition problems and datasets.
Our systematic study on the meta-parameters (e.g., image resolution, depth, training data diversity)
of DCN also provides better knowledge about how to configure the networks for future researches.

The rest of the paper is organized as follows.
In Section~\ref{sec:convnet}, we review the related works of DCN.
In Section~\ref{sec:adaptation}, we describe the proposed method for transfer learning.
In Section~\ref{sec:dataset}, we describe the datasets we used in this work and their properties,
and we summarize the DCN architectures we used in Section~\ref{sec:architecture}.
Our preliminary studies on the DCN architectures are in Section~\ref{sec:configuration}.
Experiment results for transfer learning are in Section~\ref{sec:experiment}.
Finally, we summarize the work in Section~\ref{sec:conclusion}.

\section{Deep Convolution Network for Video Recognition}
\label{sec:convnet}

In this section, we review the basic concept of convolution neural network.
We show that convolution is a specialized regularization of standard neural network 
that reduces the number of learnable parameters in the model based on prior knowledge in vision.
We then discuss previous works on evaluating DCN architectures and applying DCN for video recognition.

\subsection{Convolution neural network}
\label{sub:convolution_neural_network}
The main difficulty of applying Deep Neural Network is that it is hard to train,
due to the complexity introduced by the depth and the large number of learnable parameters.
\cite{hinton:2006fast} proposed a greedy layerwise pre-training to overcome the problem.
Pre-training is an unsupervised learning process that learns the hidden layers one-by-one,
where each layer is learned by either maximizing the input likelihood or minimizing reconstruction error.
The entire network is then fine-tuned with the labeled training data.
Pre-training is believed to learn a better initialization that captures important patterns in the data, 
which facilitates the following supervised learning and leads to better generalizability \cite{Erhan:2010jmlr}.
The unsupervisedly learned networks can be used for transfer learning,
as shown in \cite{lee:2009icml,glorot:2011icml}.

In computer vision, the most popular Deep Learning architecture is DCN \cite{LeCun:1998}.
It can be viewed as a specialized Multi-Layer Perceptron (MLP) with a manually crafted architecture and regularization.
MLP can be formulated as a series of affine transform followed by non-linear dimension-wise transform:
\begin{equation}
\label{eq:perceptron}
  h^{l} = 
  \begin{cases}
  g(\mathbf{W}^{(l)} \mathbf{x}) & \text{if layer } l = 1\\
  g(\mathbf{W}^{(l)} f^{(l-1)}) & \text{if layer } l > 1.
  \end{cases}
\end{equation}
Given the activation function $g$ and loss function, the learnable parameters $\mathbf{W}^{(l)}$ are learned by gradient descent.
One significant problem of applying MLP directly to general images is that the image signal usually contains tens of thousands of dimensions,
which leads to extremely large affine transform matrices $\mathbf{W}$.
Without clever learning process and regularization, the model will be extremely prone to overfitting.

\begin{figure}
\centering
\includegraphics[width=\linewidth]{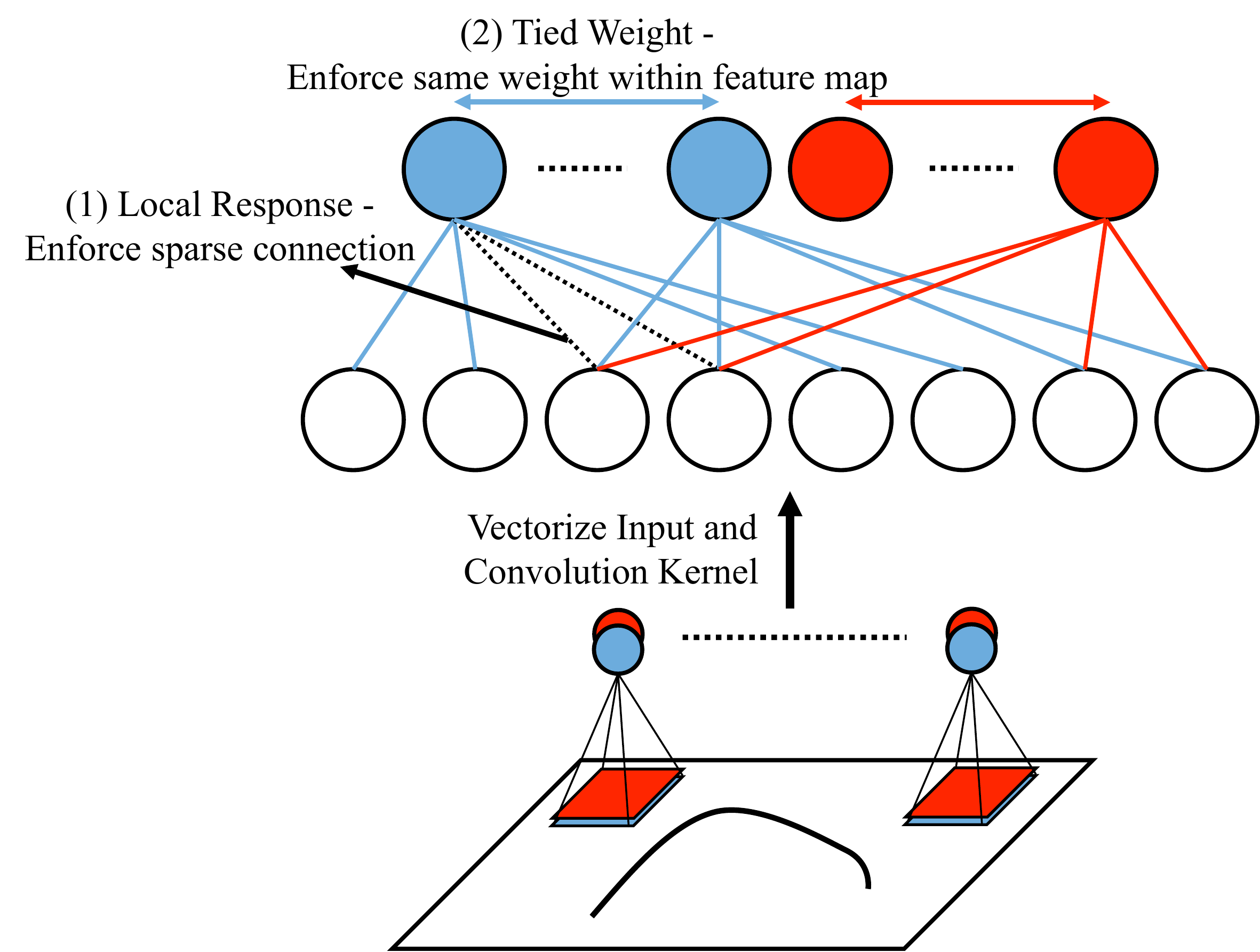}
\caption{\label{fig:convnet} Convolution illustration.
Convolution network is essentially a MLP with two additional constraints: (1) Local response and (2) Tied weight.
Local response enforces sparse connections between layers using the local receptive field heuristic
and the tied weight constraint enforces the weight within the feature map to be the same.
The two constraints reduce the number of learnable parameters and avoid overfitting.
}
\end{figure}

To avoid overfitting, convolution was introduced as a regularization on $\mathbf{W}$ 
based on the prior knowledge on the human visual system and the image signal \cite{LeCun:1998,lee:2007nips}.
Convolution is usually formulated as
\begin{equation}
  h^{l}_{k,i,j} = g(\sum\limits_{k'}\sum\limits^{N_W}_{r,s=1}\mathbf{\tilde{W}}^{l,k}_{k',r,s}h^{l-1}_{k',i-\frac{N_W}{2}+r,j-\frac{N_W}{2}+s}),
\end{equation}
where both the input and output are three dimensional tensors, and $\tilde{W}^{l,k}$ stands for the $k$-th convolution kernel in the $l$-th layer.
The width of the kernel is denoted by $N_W$.
It can be reformulated as
\begin{equation}
  h^{l}_{k,i,j} = g(\sum\limits_{k'}\sum\limits^{N_H}_{i',j'=1}\mathbf{W}^{l,k,i,j}_{k',i',j'}h^{l-1}_{k',i',j'}),
\end{equation}
where $N_H$ stands for the width of $h^{l-1}$ (assume squared input).
The convolution kernel $\tilde{W}^{l,k}$ is redefined over the entire input space at each position $(i,j)$ as $W^{l,k,i,j}$
with the following two constraints:
\begin{equation}
\label{eq:local_response}
\mathbf{W}^{l,k,i,j}_{k',i',j'}=0 \text{, if } \|i-i'\|>\frac{N_W}{2} \text{ or } \|j-j'\|>\frac{N_W}{2}
\end{equation} 
and
\begin{equation}
\label{eq:tied_weight}
\mathbf{W}^{l,k,i,j}_{k',i',j'}=\mathbf{W}^{l,k,i+r,j+r}_{k',i'+r,j'+s} \forall r,s.
\end{equation}
By vectorizing $h^{l}_{k,i,j}$ over $i,j,k$ and $\mathbf{W}^{l,k,i,j}_{k',i',j'}$ over $i,j,k$ and $i',j',k'$ respectively, 
$\mathbf{W}^{l,k,i,j}_{k',i',j'}$ will reduce to the two dimensional tensor $\mathbf{W}^{(l)}$ in Eq.~\ref{eq:perceptron},
as shown in Fig.~\ref{fig:convnet},
with two additional constraints: 
\textsl{local response} as in Eq.~\ref{eq:local_response} and \textsl{tied weight} as in Eq.~\ref{eq:tied_weight}.
The \textsl{local response} constraint enforces that only a small part of $\mathbf{W}^{l,k,i,j}$ can be non-zero,
which means that $h^{l}_{k,i,j}$ can only depend on a small fraction of $h^{l-1}$ while it can have arbitrary dependency in MLP.
This reduces the number of learnable parameters with the heuristic that local patterns are important for both visual recognition and human vision system.
\textsl{Tied weight} further reduces the number of learnable parameters by 
enforcing kernels $\mathbf{W}^{l,k,i,j}$ with same $l,k$ to share the same value.

Despite the convolution regularization, DCNs still have a large amount of learnable parameters and are still prone to overfitting.
While unsupervised pre-training is popular in DBN to improve the generalizability, 
DCN as a fully supervised learning algorithm does not utilize similar learning techniques.
Therefore, a large training set with high diversity is necessary to learn a DCN, yet such datasets are not easily obtainable in the past.
Also, the computation cost of convolution made training on moderate resolution images (200x200, etc.) a formidable task until very recently.
The most impressive breakthrough of DCN on visual recognition comes from its superior performance on the ImageNet dataset,
where \cite{Krizhevsky:2012nips} and \cite{Le:2012icml} 
independently report a significant performance improvement over traditional image features.
While the network architecture used by each group is significantly different from each other,
the key factors for the success of both groups are the extremely large training set as well as the parallel acceleration that makes learning possible.
This partially explains the resurge of Neural Networks, which had been developed long before they received much attentions:
it is only recently that such large training sets as well as computation power have become available for the network to be learnable.

Except the learnable layers mentioned above, 
most DCN models also include some ``fixed layers'' that were manually designed to inject desirable properties to the model.
The most common fixed layer is the pooling layer.
A pooling layer combines several outputs from the previous layer, usually a local 2x2 or 3x3 patch, 
and generate one single output using some statistics over the input such as maximum or average.
The operation is similar to convolution in the sense that it is computed locally over the entire input,
except it use a different activation function $g$ and does not adapt with data.
The pooling layer is introduced for two main purposes.
First, by taking local statistics, it introduce certain degree of translation invariance to the model.
Invariance properties are considered important for visual features, but they are hard to realize in convolution neural network.
Although it can handle only small pixel shift depending on the size of receptive field,
the pooling layer will generate the same output for inputs differing with small translation and therefore realize translation invariance.
Second, the pooling layer can reduce the computational cost of DCN.
Consider a 2x2 pooling without overlap. 
The output size will reduce by a factor of 4 after the convolution layer.
Therefore, the following layers can operate on a smaller 2D maps,
which reduce both the computation and memory consumption.
There exist other popular fixed layers such as response normalization,
but they are not as popular as pooling and may sometime harm the performance.

\subsection{Network configuration}
\label{sub:network_configuration}
Convolution neural networks have a large number of meta-parameters,
including the network architecture, initialization, etc.
To specify the network architecture, we have to decide the depth of the network, 
the number of hidden units or convolution kernels in each layer, the receptive field of each layer and many other details.
It is known that meta-parameters have a great impact on the network performance \cite{goodfellow2013arxiv,simonyan2014arxiv,zeiler2014eccv}.
However, due to the large number of possible configurations and the computational cost,
it is impossible to perform grid search for the optimal configuration in most situations.
There exist several popular network architectures in the research community \cite{Krizhevsky:2012nips,szegedy2014arxiv,simonyan2014arxiv},
and many previous works use these configuration without further optimization \cite{Girshick:2014cvpr, Oquab:2014cvpr},
Although being successful in many cases, most of the architectures are proposed without explanation or justification of the choice.
This leads to the lack of knowledge about how to improve the configuration for specific task,
which can be expected to improve the performance of the model.

Some previous works have reported more details about their configuration design.
\cite{szegedy2014arxiv} provides high level arguments for their motivation for increasing network depth and connection design,
but they do not provide experiment evidences for the particular choice of architecture.
\cite{goodfellow2013arxiv,simonyan2014arxiv} provide systematic experiment results to justify their choice of network architectures,
but both of them focus only on the depth of the network while there exist many other factors in network architecture.
\cite{zeiler2014eccv} optimizes the step size and receptive field of the convolution kernel in first layer,
based on the model proposed in \cite{Krizhevsky:2012nips}.
However, the update is based on empirical analysis of the visualization result 
and is hard to generalize to other data or higher layers in the network.
\cite{chatfield2014arxiv} evaluates various configurations of DCN on image recognition including network architecture,
but they only test three different architectures which provide limited information on the effect of meta-parameters.
Compared to all existing works, our evaluation on network architecture is more complete and systematic.
Also, we are the first to study the effect of input image resolution on DCN,
which may help to significantly reduce the computational cost in certain scenarios.

\subsection{Deep convolution network for video recognition}
\label{sub:deep_convolution_network_for_video_recognition}
Motivated by the outstanding performance of applying DCN on static image recognition,
there is a growing interest of applying DCN on video recognition.
An intuitive extension of DCN to video is to perform convolution over a spatio-temporal volume,
so the 3-dimensional convolution kernel can capture short term local motion instead of only spatial motion in original 2-dimensional DCN.
This extension has been shown successful on various action recognition benchmarks \cite{le2011cvpr,ji2013pami,pei2015visualcomputer},
but it does not perform well on recognizing more complex events \cite{Karpathy:2014cvpr}.
\cite{Karpathy:2014cvpr} proposes a more general framework for fusing multiple frames,
and they evaluate different fusion strategies on million scale video dataset for recognizing sports event.
Their results suggest that single-frame DCN performs similarly to more complex models.

Instead of extending convolution to temporal domain, a more common approach is to build the video recognition model on top of the frame-wise DCN.
\cite{xu2014arxiv,zha2015arxiv} exploits different pooling methods to combine frame features extracted by static image DCN.
They show that the pooling method has significant impacts on video recognition performance, 
and by properly pooling the frame-wise features, static image DCN can significantly outperform more complex convolution models.
\cite{jiang2015arxiv} combines DCN features with other hand crafted features using another neural network,
where the video DCN representation is the average of frame-wise features extracted by static image DCN.
\cite{simonyan2014nips} proposes a late fusion approach to combine local motion and appearance information.
The local motion of each frame is first captured by optical flow and then processed by a frame-based convolution network,
instead of convolving over spatio-temporal volumes.
One problem of 3-dimensional convolution is that it captures only temporally local motion, 
while event recognition usually requires the long term dependency of high level semantic concepts.
\cite{donahue2014arxiv,wu2015arxiv} use Long Short Term Memory network on top of the frame-wise DCN feature to model the long-term dependency of 
high level semantic concepts. They show that the recurrent neural network can further boost the performance of frame-wise DCN.
While our method is also based on pooling frame-wise DCN results, 
in contrast to most existing works that focus on improving the pooling method, 
we aim at improving the DCN model itself.
Therefore, it can be combined with any of the static image DCN based models mentioned in the previous paragraph.

\section{Transfer Learning with Deep Convolution Network}
\label{sec:adaptation}
In this section, we describe the transfer learning approaches we apply to utilize image information in video recognition.
Transfer learning helps to learn a more generalizable DCN.
This is important because DCNs are prone to overfitting, especially when only scarce training data is available.
While increasing training data helps to solve the problem, there are cases where collecting new data with complete ground truth is difficult.
Transfer learning solves the problem by using labeled data from other domains 
where a large number of training data is available to improve the network.

The goal is similar to the pre-training process in Deep Belief Network (DBN) and Stacked Auto-Encoder (SAE) in the sense that 
it improves the generalizability by learning a better intermediate representation \cite{Erhan:2010jmlr}.
A good representation does not necessarily optimize the loss during training, 
which is done by supervised backpropagation in Deep Neural Network;
instead, it should capture important patterns that are general to all data.
While DBN and SAE achieve this by performing unsupervised pre-training before supervised training,
we learn the representations from other domains and then optimize the representation through transfer learning.

\begin{figure}[t]
\centering
\includegraphics[width=0.8\linewidth]{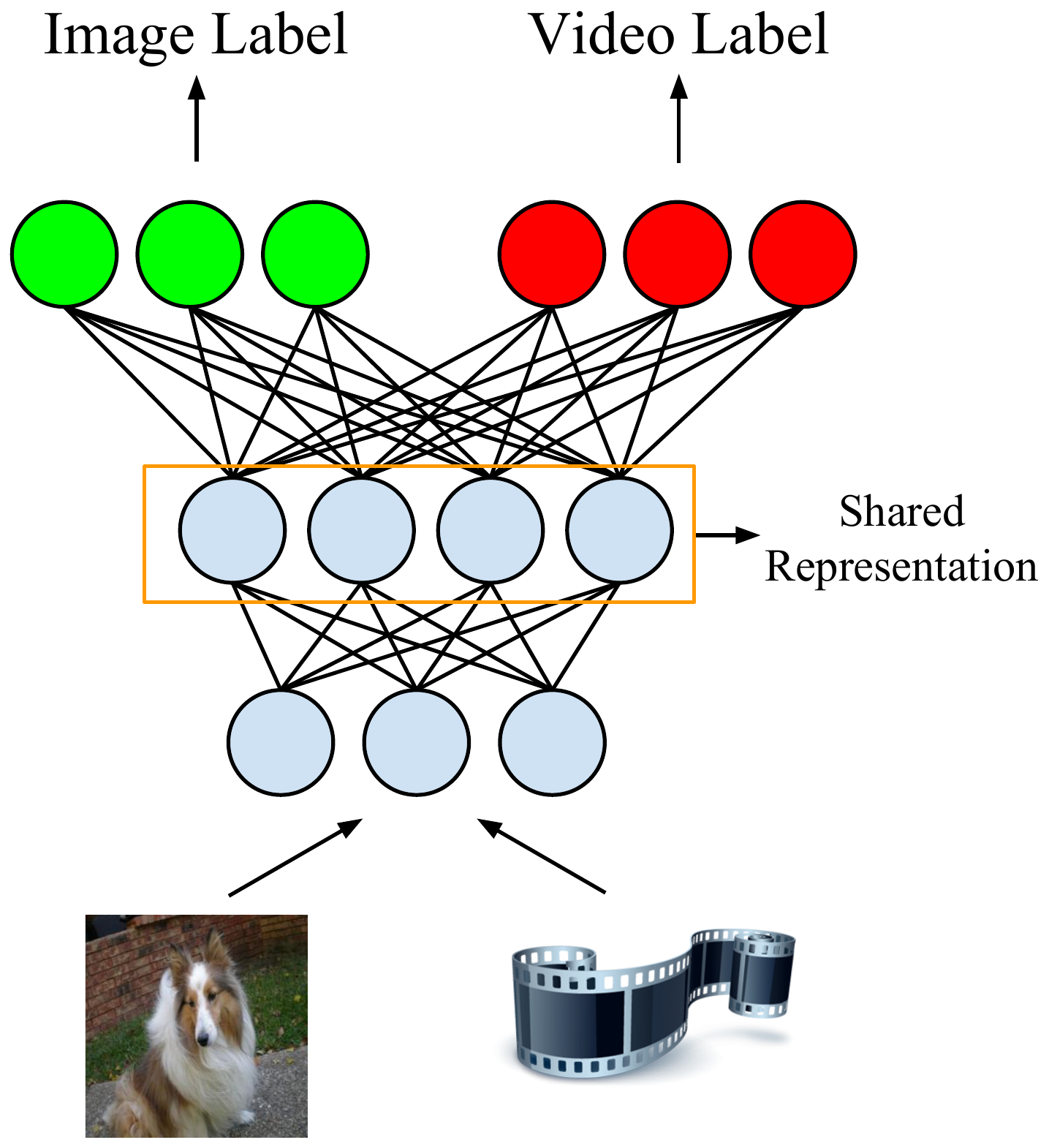}
\caption{\label{fig:shared_feature} Augment training set for transfer learning.
The DCN is trained with both images and video frames simultaneously. 
Because the middle layers are shared by all output units,
the internal representation benefits from both the image and video domains and learns a more robust network.
This helps to avoid overfitting and learn general visual patterns in natural images.
}
\end{figure}

\subsection{Augment training data}

The first approach for transfer learning is to augment video data with independent image data.
Or equivalently, we train a DCN that simultaneously recognizes images from the image datasets and frames from the video dataset,
as illustrated in Fig.~\ref{fig:shared_feature}.
This is made possible since the intermediate layers are shared by all output units in Neural Networks 
and may benefit from the training data of other classes.
If the lower layers can learn general visual patterns that are shared across different datasets,
the additional classes and training data can help to learn better low and middle level features and avoid overfitting.

\subsection{Transfer mid-level features}

\begin{figure}[t]
\centering
\includegraphics[width=\linewidth]{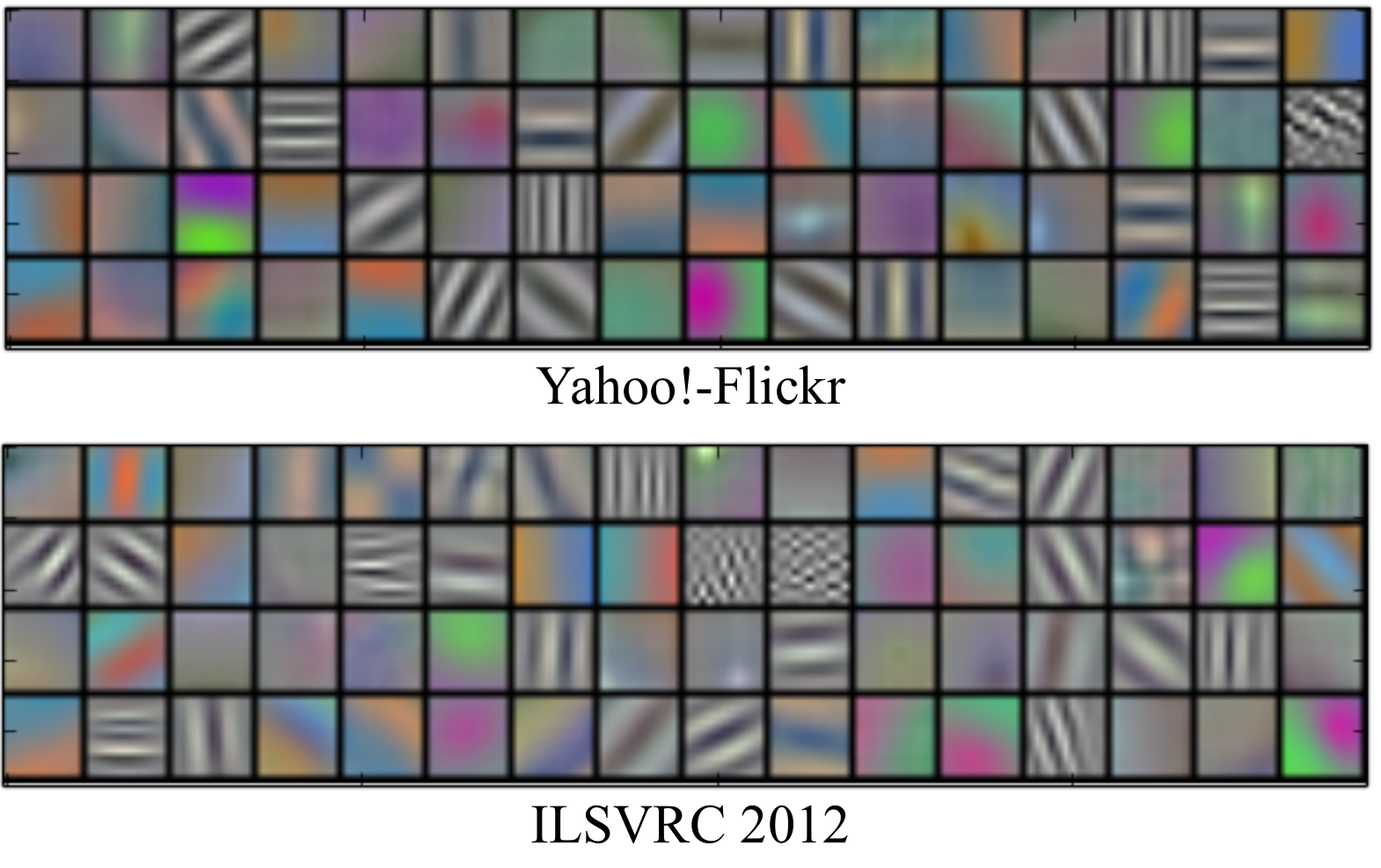}
\caption{\label{fig:imagekernels} First convolution layer kernels learned from \textsl{Yahoo!-Flickr} and \textsl{ILSVRC2012} datasets.
Despite being learned from non-overlapping datasets, some kernels are visually similar,
which supports the assumption that some important image patterns are shared across different datasets provided enough training data.
}
\end{figure}

The second approach for transfer learning is to transfer the learned feature from images to videos
by initializing the learnable parameters of DCNs using the network pre-trained on image domain.
The approaches can also be considered as a supervised pre-training,
which is analogous to the unsupervised pre-training in DBN or SAE.
The network is then fine-tuned using the target dataset to optimize the features for the target domain.
This approach is motivated by the fact that image features rely on important visual patterns that are shared across all natural images,
so they can be used to characterize images outside the training set.
Because the convolution kernels in DCN learn important low level visual patterns in natural images \cite{Le:2012icml},
these convolution kernels serve as the low level features used in traditional visual recognition,
and they may also be similar across all natural images and datasets.
This can be seen in Fig.~\ref{fig:imagekernels},
where some of the first layer kernels are very similar even if they are learned from two non-overlapping datasets.
Therefore, the network parameters learned from one dataset may be useful for another dataset.

The fine-tuning step is performed to further optimize the feature.
This is especially important for the higher layers in DCN, such as the fully connected layer,
because these layers capture more complex patterns \cite{Le:2012icml} that may not generalize well to other domains.
For example, while lines and corners are common to all natural images,
a pattern of face will appear in only specific domains and may not be useful in all problems.
Because training neural networks as an optimization problem is non-convex, 
different initial values of the learnable parameters will lead to different networks.
In fact, it is known that good initialization will lead to better performance \cite{sutskever:2013icml}.
Therefore, by initializing the learnable parameters with visual patterns learned from independent image corpus,
we should be able to learn more generalizable networks, as suggested by the pre-training process.

While previous researches focus on unsupervised pre-training for better representation learning, 
several parallel works have utilize the supervised pre-training approaches to address the same lack-of-training-data problem \cite{donahue2013decaf,Oquab:2014cvpr,Girshick:2014cvpr,Karpathy:2014cvpr}.
In most of these works, DCNs learned on the \textsl{ILSVRC} dataset are applied on other static visual recognition problems 
such as object recognition on the \textsl{Pascal VOC} dataset.
Our work is complementary to them in that we show the same approach can be well generalized in both source and target domain.
The dataset we used for supervised pre-training is a weakly labeled dataset that uses image tags as ground truth,
and the DCNs are then applied to Youtube videos rather than static images.
Also, some of our discoveries and conclusions are consistent to those in \cite{Girshick:2014cvpr,Karpathy:2014cvpr} as described below.

\section{Dataset}
\label{sec:dataset}
\begin{figure*}
\centering
\includegraphics[width=\textwidth]{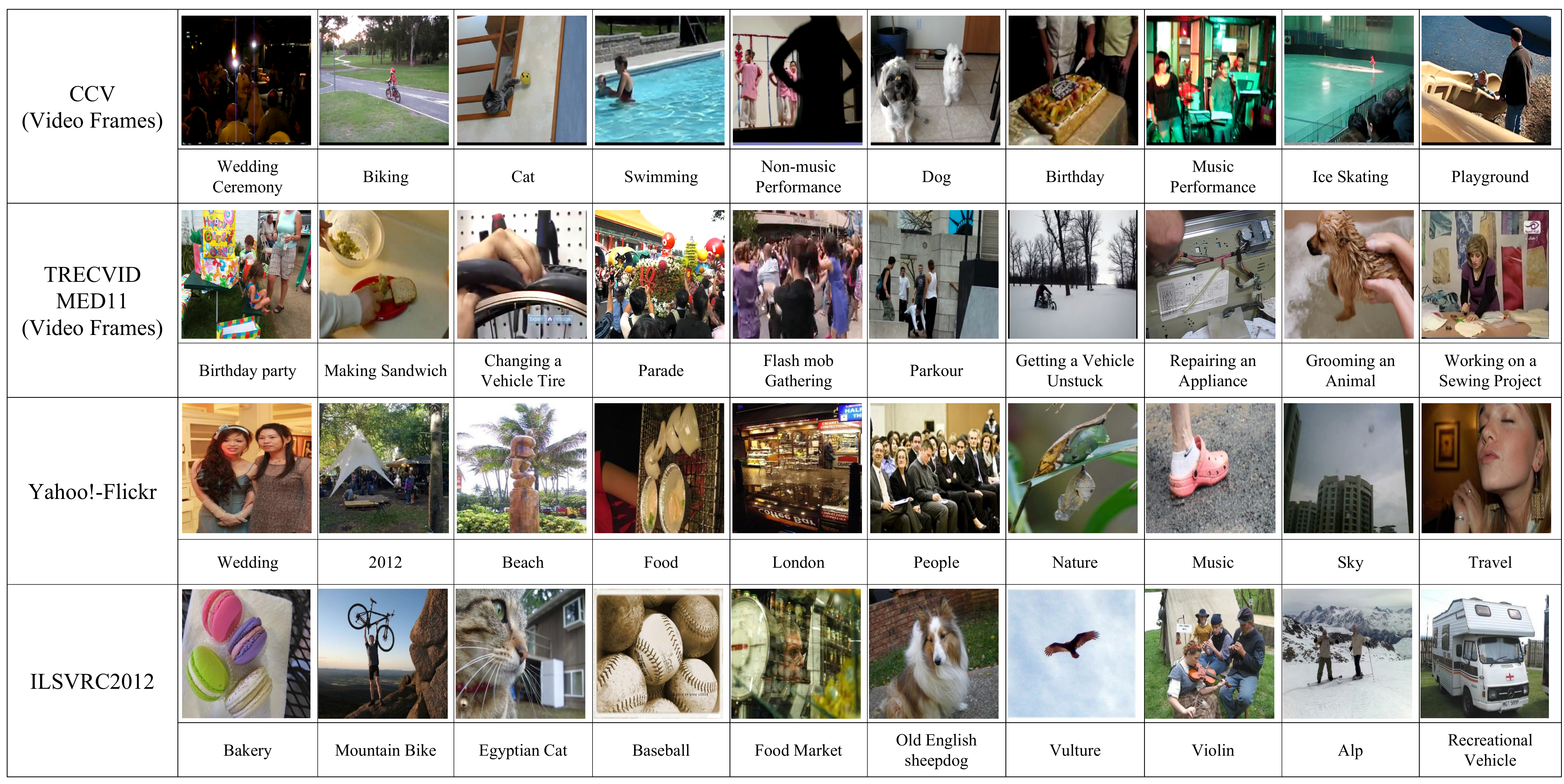}
\caption{\label{fig:example_image} Example images from the dataset used in this work.
These images show some important properties of these datasets.
\textsl{ILSVRC2012} is a dataset for object recognition, where the object may occupy only a small part of the image.
All the images are manually annotated, so the label is reliable.
The large number of classes is the main challenge of the dataset.
\textsl{Yahoo!-Flickr} is for tag prediction, 
and the tags are mostly high level description (e.g., scenes) of the image which depends on the entire image or meta information.
The tags are collected from the internet without further annotation or verification, so the label is known to be noisy.
Because it contains only 10 categories, the intra-class diversity is also larger than most image recognition benchmarks.
\textsl{CCV} and \textsl{TRECVID MED11} are the video recognition datasets, 
where both of them are labeled with high level semantic concepts such as scene or events.
}
\end{figure*}

In this section, we describe the datasets we used and discuss their properties.
We use 4 datasets in our experiments, 2 static image datasets and 2 video datasets.
The image datasets are used to compare different network architectures and perform transfer learning to boost video recognition DCN.
While both of them are million scale image datasets,
they have certain fundamental differences including label quality, number of categories and the recognition targets.
These differences stem from the data collection process and dataset design.
Because our goal is to recognize high level semantic concepts from the video,
the video datasets are labeled with either semantic concepts or complex events that involve the interaction of various semantic concepts.
Both of them are popular in previous video recognition researches.
We also show the example images in Fig.~\ref{fig:example_image}.

\textbf{ILSVRC2012.} ILSVRC2012 \cite{ILSVRC2012} is an 1k classes subset of ImageNet used in ImageNet Large Scale Visual Recognition Challenge.
While the 1k classes may be either internal or leaf nodes of the ImageNet ontology, they are guaranteed to be non-overlapping.
Each class correspond to an object, and an image is considered positive to the class if the object appears in the it.
The entire dataset comprises of 1.5 million manually labeled images, 
where 1.2 million of them are used as the training set and 50,000 images are used as the validation set;
we report the performance on the validation set.
The dataset is characterized by its large number of classes and is often used to evaluate visual recognition system with large semantic space.
Because all the images are manually annotated, the label is considered precise and reliable.

\textbf{Yahoo!-Flickr.} \textsl{Yahoo!-Flickr} contains images with 10 different classes, 
where the classes are defined by popular Flickr tags in and the ground truths are obtained directly from Flickr without further annotation or verification.
Therefore, it contains only weak supervision, 
and the label is known to be noisy because of user tagging behavior \cite{Su:2013mmgc}.
Each class contains 150k images for training and 50k images for testing.
\textsl{Yahoo!-Flickr} dataset is characterized by its large intra-class variations,
where the images in each class may be visually very diverse or even share no visual similarity at all.
Also, the classes usually correspond to more vague concepts such as ``Sky'' instead of some concrete objects.

\textbf{CCV.} 
Columbia Consumer Video (CCV) Database \cite{jiang:2011icmr} is a video dataset that targets on unconstrained consumer video content analysis.
The dataset contains 9,317 manually labeled Youtube videos with 20 semantic categories, where each video may belong to arbitrary number of categories.
The 9,317 videos are divided into an equal-sized train and test set, 
and the general evaluation protocol adopts binary relevance multi-label classification on the 20 semantic categories with mean average precision (MAP) criterion.

\textbf{TRECVID MED11.}
TRECVID MED is the dataset used in the multimedia event detection task in annual TRECVID activity.
It consists of consumer generated videos published on video sharing sites and is collected and annotated by the Linguistic Data Consortium.
We use the 2011 release, which contains 13,115 training and 32,061 test videos.
15 event classes were defined in this release, with 5 of them annotated only in the training set for system development.
Each video contains at most one out of the 15 events, annotated at video level.
We use all 15 event classes for training and use the videos with event label following \cite{izadinia2012eccv}.

To learn a DCN for the video dataset, 
we first perform uniform sampling on the training set to extract image frames for training,
and we treat each frame as an independent sample.
In the testing phase, we perform recognition on the keyframes independently and use late fusion to combine the confidence score of different frames.

\section{Network Architecture}
\label{sec:architecture}
In this section, we describe the network architecture and other parameters we used for our experiments.
We will refer to these architectures based on the input image resolution and number of convolution layers in the following sections.

\textbf{2-convolution-layer network.} 
For networks with two convolution layers, we adopt an architecture similar to that of LeNet5 \cite{LeCun:1998}.
The network is composed of 5 layers; the first two layers are convolution layers, and the third and fourth layer are fully connected layers.
The last layer is a classification layer.
The structure of convolution layers depend on the image resolution.
For 256x256 and 128x128 input image resolution, the first convolution layer has 64 kernels of size 11x11x3 with a stride of 4 pixels.
The second convolution layer has 128 kernels of size 5x5x64. 
Max pooling with 2x2 region is performed after each convolution layer.
The third layer, or the first fully connected layer has 4,096 neurons;
the second fully connected layer has 1,024 neurons for \textsl{Yahoo!-Flickr} dataset and 2,048 for \textsl{ILSVRC2012}.
The structure of fully connected layers are kept the same over all input image resolution and network architecture.
Given a 32x32 input image resolution,
the first convolution layer has 64 kernels of size 5x5x3 and the second convolution layer has 128 kernels of size 5x5x64.
For a 64x64 input image resolution, 
the kernel sizes are the same with the exception that we use 32 kernels for the first layer and 64 for the second layer.
If not mentioned specifically, all convolution layers have stride 1 and are followed by 2x2 max pooling,
and Rectified Linear Unit\footnotemark (ReLU) activation is used in all the network layers.
\footnotetext{
ReLU is an approximation of sigmoid function: \\$f(x)=x*\mathbb{I}(x>0)$, where $\mathbb{I}()$ is the indicator function.
}

\textbf{3-convolution-layer network.}
The 3-convolution-layer network is modified from 2-convolution-layer network, with the following differences:
for 256x256 and 128x128 input image resolution, the first convolution layer now uses kernel of size 7x7x3 with a stride of 3;
the third convolution layer has 320 kernels of size 3x3x128, and no max pooling is performed on the output of the third convolution layer.
The second convolution layer and fully connected layers are the same as the 2-convolution-layer network with different input size.
For 64x64 input image resolution, the first convolution layer consists of 32 kernels of size 5x5x3, 
the second layer consists of 64 kernels of size 5x5x32, and the third layer has 120 kernels of size 3x3x64;
max pooling is performed after the first and second convolution layer.
For 32x32 input image resolution, the convolution layer structures are similar to that of 64x64 input image resolution,
except we perform max pooling only after the first layer, otherwise the third layer will be too small.

\textbf{4-convolution-layer network.}
The 4-convolution-layer network is similar to the architecture used in \cite{Krizhevsky:2012nips} with some simplification.
The first convolution is the same as that in the 2-convolution-layer network.
The second convolution layer has 128 kernels of size 5x5x64.
The third convolution layer has 192 kernels of size 3x3x128, and the fourth convolution layer has 192 kernels of size 3x3x192.
Note that we do not perform response normalization and grouping on the convolution kernels as in \cite{Krizhevsky:2012nips},
and our max pooling is performed over 2x2 regions without overlapping.

\textbf{5-convolution-layer network.}
The 5-convolution-layer network is based on the architecture proposed in \cite{Krizhevsky:2012nips} with the following modification.
First, the number of hidden units in fully connected layer is reduced, as described in 2-convolution-layer network architecture.
Second, we remove the response normalization layer in the network following the 4-convolution-layer architecture.
It is worth noting that the 5-convolution-layer network has more convolution kernel than other architectures we studied:
the first layer has $50\%$ more convolution kernels, while the second to fourth layers have twice the number of kernels.
Therefore, it is expect to have better performance.
We choose the setting to make the result comparable with existing works.

The Caffe package \cite{Jia:2013caffe} is used for learning the networks.
The learning process is similar to that in \cite{Krizhevsky:2012nips}, where we crop and mirror the images to produce more training samples.
Images of size 256x256 are cropped to a size of 227x227, 128x128 to 116x116, 64x64 to 56x56 and 32x32 to 28x28.
The same dropout ratio, initial learning rate and momentum as in \cite{Krizhevsky:2012nips} are used.

\section{Experiment -- Network Configuration Selection}
\label{sec:configuration}

Instead of performing computation intensive grid search on meta-parameters for every new problem,
we believe that by studying the correlation between meta-parameters and the network performance on a wide range of datasets,
we can learn a heuristic that gives us a reasonable network configuration based on the target problem.
Therefore, we perform extensive experiments on image datasets to better understand the properties of DCN.
We set the configuration of frame-based video recognizer based on these knowledge without performing cross validation on the target video datasets.
In particular, we study the image resolution for input image and the required depth of the network,
which have the greatest effect on computation cost of DCN.
We also study how to sample the frames from video,
which manifests the fact that a large training set with high diversity is necessary for training DCN.

\begin{table}[t]
  \centering
  \caption{Performance comparison of different input image resolutions on the \textsl{ILSVRC2012} and \textsl{Yahoo!-Flickr} dataset.
  For \textsl{ILSVRC2012}, decreasing the image resolution significantly degrades the performance,
  while the performance on \textsl{Yahoo!-Flickr} is less sensitive to input image resolution.
  The results reflect the fact that \textsl{ILSVRC2012} is an object level recognition dataset, 
  where the object may take only a small part of the image and decreasing the image resolution will make the object unrecognizable.
  \textsl{Yahoo!-Flickr}, on the other hand, is a scene level recognition dataset, and the category depends on the entire image.
  Therefore, it is still recognizable using thumbnail images.
  Nevertheless, high resolution images always yield better performance.
  Also note the poor performance of 3-convolution-layer network with 32x32 input image resolution,
  which indicates the limit in depth imposed by input image resolution (due to max-pooling).
  Therefore, we use 256x256 input image resolution for following experiments.
  }
  \vspace{5pt}
  \begin{tabulary}{\textwidth}{C|C||C|C||C|C}
    \hline
    \multirow{2}{*}{Res.}     & \multirow{2}{*}{Depth}  & \multicolumn{2}{c||}{ILSVRC2012}  & \multicolumn{2}{c}{Yahoo!-Flickr}\\
    \cline{3-6}
                    &               & Top-1       & Top-5     & Accuracy    & MAP\\
    \hline
    \hline
    \multirow{2}{*}{32x32}    & 2-conv.           & 0.26        & 0.47      & 0.48      & 0.46\\
    \cline{2-6}
                  & 3-conv.           & 0.23        & 0.43      & 0.45      & 0.43\\
    \hline
    \multirow{2}{*}{64x64}    & 2-conv.           & 0.31        & 0.55      & 0.51      & 0.50\\
    \cline{2-6}
                  & 3-conv.           & 0.31        & 0.55      & 0.49      & 0.47\\
    \hline
    \multirow{2}{*}{128x128}  & 2-conv.           & 0.31        & 0.54      & 0.50      & 0.50\\
    \cline{2-6}
                  & 3-conv.           & 0.39        & 0.63      & 0.51      & 0.51\\
    \hline
    \multirow{2}{*}{256x256}  & 2-conv.           & 0.40        & 0.64      & 0.54      & 0.53\\
    \cline{2-6}
                  & 3-conv.           & 0.46        & 0.71      & 0.56      & 0.56\\
    \hline
  \end{tabulary}
  \label{tab:resolution}
\end{table}

\subsection{Image Resolution}

Early experiments of DCN are mostly based on datasets with very low image resolution.
While it is claimed that these thumbnail images are still human recognizable \cite{Torralba:2008},
it is obvious that images with higher resolution contain more detailed information that may be helpful for visual recognition.
Therefore, state-of-the-art DCN models usually operate on images with higher resolutions (256x256, etc.).
But higher resolution also implies higher computational cost. 
In DCN, the complexity grows roughly quadratically as the image resolution increases.
To minimize the computational cost,
we try to investigate whether high resolution images are necessary for DCN in general visual recognition.

We resize the images of \textsl{Yahoo!-Flickr} and \textsl{ILSVRC2012} into four different resolutions ranging from 256x256 to 32x32
and then train DCNs with either 2 or 3 convolution layers on each resolution.
The results are in Table~\ref{tab:resolution}, 
where the two datasets show different responses with respect to image resolution.
For the \textsl{ILSVRC2012} dataset, the performance is very sensitive to the image resolution,
and we can consistently obtain $10\%\sim15\%$ relative improvement by doubling the resolution.
Also note that 3-convolution-layer network has worse performance than 2-convolution-layer network with 32x32 input image resolution,
which implies the limit on depth imposed by image resolution.
In fact, we have to abandon max pooling after the second convolution layer of the network to have a large enough hidden layer,
otherwise the network will be nearly unlearnable and have extremely bad performance.
The \textsl{Yahoo!-Flickr} dataset, on the other hand, shows only moderate performance degradation when reducing the resolution,
and we can achieve reasonable performance even with 32x32 thumbnail images.

The difference of the two datasets stems from the fact that they are designed for different purpose and have different properties.
In particular, while the \textsl{ILSVRC2012} dataset is designed for object recognition where the object may be present in only a small part of the image,
the \textsl{Yahoo!-Flickr} dataset is designed for tag prediction, where the tags are mostly high level concepts that describe the entire image.
The results indicate that we may use smaller images without significant loss of performance yet reduce the computational cost quadratically.
Nevertheless, increasing the resolution consistently yields better performance and enables the usage of deeper networks,
so we use 256x256 resolution in following experiments.

\subsection{Depth of Architecture}

In this section, we compare the performance of DCNs with different numbers of convolution layers.
Since adding layers in the network significantly increases the computational cost, 
we would like to use networks as shallow as possible if the additional layers have no or even negative contributions on the performance.

\begin{table}[t]
  \centering
  \caption{Performance comparison of DCNs with different depths on the \textsl{Yahoo!-Flickr} and \textsl{ILSVRC2012} dataset.
  For the \textsl{Yahoo!-Flickr} dataset, 3-convolution-layer network achieves the best performance, 
  but the performance gain over 2-convolution-layer network is minor.
  The 4-convolution network has much worse performance, which shows that deeper network may even degrade the performance.
  For the \textsl{ILSVRC2012} dataset, on the other hand, the additional layers significantly improve the performance,
  and the performance grows monotonically with the number of layers.
  }
  \vspace{5pt}
  \begin{tabulary}{\linewidth}{L||C|C|C|C}
    \hline
    \multicolumn{5}{c}{Yahoo!-Flickr}\\
    \hline
    Depth & 2-layers & 3-layers & 4-layers & 5-layers\\
    \hline
    Accuracy & 0.535 & 0.560 & 0.491 & 0.531 \\
    \hline
    MAP & 0.534 & 0.559 & 0.478 & 0.524 \\
    \hline
    \hline
    \multicolumn{5}{c}{ILSVRC2012}\\
    \hline
    Depth & 2-layers & 3-layers & 4-layers & 5-layers\\
    \hline
    Top-1 & 0.40 & 0.46 & 0.51 & 0.55 \\
    \hline
    Top-5 & 0.64 & 0.71 & 0.75 & 0.79 \\
    \hline
  \end{tabulary}
  \label{tab:yahoo_depth}
\end{table}

To evaluate the effect of different depths on the performance,
we learn convolution networks with 2$\sim$5 convolution layers.
The results are in Table~\ref{tab:yahoo_depth}.
Again, the two datasets show difference response to the number of convolution layers.
In \textsl{Yahoo!-Flickr},
3-convolution-layer network turns out to have the best performance,
although the performance gain of the 3rd convolution layer is relative minor.
Note the superior performance of 5-convolution-layer network over 4-convolution-layer one benefits not only from the additional layer
but also from the additional convolution kernel in each layer.
The performance gain of additional convolution layers is more significant in \textsl{ILSVRC2012},
which achieves 10$\%$ relative improvement in terms of Top-1 result.
Also, the performance grows monotonically with the number of layers.
Different from many previous works,
we find the performance of DCN does not always grow when the network becomes deeper but is data-dependent.

\subsection{Training Data Number and Diversity}
\label{sec:sampling}

In this section,
we study the effect of increasing training set size on the \textsl{Yahoo!-Flickr} datasets.
We choose \textsl{Yahoo!-Flickr} because the study of performance dependency on training set size is part of the goal the dataset is designed for.
The \textsl{ILSVRC2012} dataset, on the other hand is not suitable for the study,
because reducing the training set may either lead to very few samples for some classes or change the ratio of different classes.

\begin{table}
  \centering
  \caption{The performance of 2-convolution-layer networks on \textsl{Yahoo!-Flickr} with different numbers of training samples and cycles.
  When only 20k samples per class are used for training,
  the network suffers from significant overfitting and has poor performance on the test set.
  The results indicate that a large training set is necessary for learning a robust DCN.
  We choose 20k samples per class for comparison because the performance of linear SVM saturates around 20k training samples per class.
  }
  \vspace{5pt}
  \begin{tabulary}{\linewidth}{L||C|C|C|C|C|C}
    \hline
    Cycles & \multicolumn{2}{c|}{5} & \multicolumn{2}{c|}{10} & \multicolumn{2}{c}{20}\\
    \hline
    Training Size & 20k & 150k & 20k & 150k & 20k & 150k\\
    \hline
    \hline
    Loss (Train) & 0.62 & 1.27 & 0.33 & 1.30 & 0.19 & 1.12\\
    \hline
    Loss (Test) & 1.65 & 1.54 & 1.67 & 1.49 & 1.72 & 1.44\\
    \hline
    Accuracy & 0.44 & 0.51 & 0.44 & 0.52 & 0.43 & 0.54\\
    \hline
  \end{tabulary}
  \label{tab:yahoo_size}
\end{table}

We train 2-convolution-layer networks on \textsl{Yahoo!-Flickr} using both 20k and 150k training samples for each classes.
We choose 20k training samples for comparison 
because previous results show that the performance of linear classifier saturates with 20k training samples per-class.
The results are shown in Table~\ref{tab:yahoo_size}, 
which clearly shows that smaller training set leads to significant overfitting and therefore poor performance.

To generate large enough training set from the video datasets,
we perform uniform sampling on video frames instead of using only keyframes.
We compare the results using 1-fps sampling and 4-fps sampling on \textsl{CCV} dataset,
which leads to roughly 400k training samples and 1.6 million training samples respectively.
Although 4-fps leads to a much larger training set, 
experiment results show that the performance using 4-fps are nearly identical with 1-fps while it takes much more storage and computation.
The reason is that 4-fps sampling results in a large dataset with small visual diversity, 
and the redundant training data are not helpful for learning.
Therefore, we use 1-fps sampling to produce the training data for the following video recognition experiments.

\section{Experiment -- Video Recognition with Transfer Learning}
\label{sec:experiment}

In this section, we examine the proposed transfer learning approach on video recognition datasets.
The entire workflow of video recognition is as follows:
we sample video frames from the training set using 1-fps uniform sampling and train a frame-wise DCN.
The label of the video is propagated to all frames extracted from the video,
despite there may be irrelevant frames.
For test videos, we extract keyframes from the video and perform frame-wise recognition.
The recognition results of all the keyframes within a video are then aggregated using average pooling to predict the category of the video.

\subsection{Augment training sets}

\begin{figure}[t]
\centering
\includegraphics[width=\linewidth]{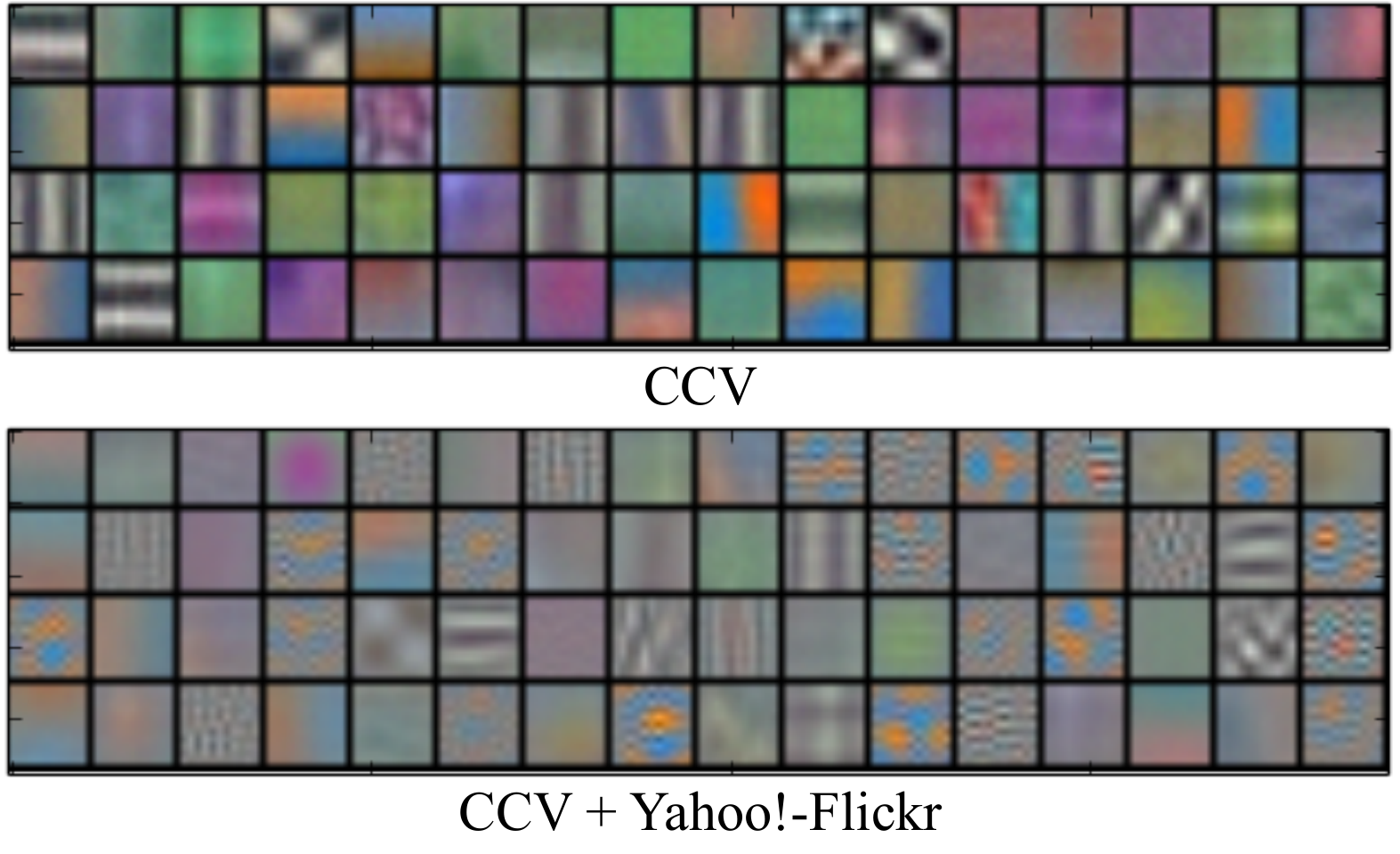}
\caption{\label{fig:videokernels} Samples of the first convolution layer kernels learned from the \textsl{CCV} dataset
and from augmenting \textsl{CCV} with \textsl{Yahoo!-Flickr}.
The kernels learned from \textsl{CCV} only show very different patterns compared with those learned from 
\textsl{Yahoo!-Flickr} or \textsl{ILSVRC2012}, where there are nearly no high frequency signals in the kernels.
By augmenting \textsl{CCV} with \textsl{Yahoo!-Flickr}, we introduce high frequency kernels into the network.
This may explain why training DCN on \textsl{CCV} only yields significant overfitting and poor performance:
it ignores informative signals.
}
\end{figure}

\begin{table*}[t]
    \def\arraystretch{1.2}
    \caption{Experiment results on the \textsl{CCV} and \textsl{TRECVID MED11} dataset.
    Depth indicates the architecture of the network, which is described in details in section~\ref{sec:architecture}.
    Initialization indicates how the learnable parameters in the network are initialized: 
    Random means random initialization, 
    while \textsl{Yahoo!-Flickr} and \textsl{ILSVRC2012} indicate initialization with the network parameters pre-trained on each dataset.
    Because the networks learn important patterns for natural images from the image datasets,
     pre-trained networks start with better intermediate representation and has better generalizability.
    Training set indicates whether we use only the video dataset or augment it with the image dataset for training.
    Because the intermediate layers are shared by all output units,
    they can learn simultaneously from the image and video domains, which leads to better performance.
    Update policy indicates which parameters are updated during training, where FC means fully connected layers and CONV means convolution layers.
    By restricting the update of convolution layers, we reduce the number of learnable parameters and avoid overfitting.
    The results show that our transfer learning approaches improve the performance of DCN.
    }

    \begin{subtable}{0.5\textwidth}
        \begin{tabu}{c|c|c|c|c}
        \hline
        Depth                       & Initialization            & Training Set                  & Update Policy & MAP \\
        \hline
        \multirow{11}{*}{2-Conv.}   & \multirow{3}{*}{Random}   & CCV                           & FC + CONV     & 0.445 \\
                                                                \cline{3-5}
                                    &                           & CCV + Yahoo!                  & FC + CONV     & 0.469 \\
                                                                \cline{3-5}
                                    &                           & CCV + ILSVRC                  & FC + CONV     & 0.469 \\
                                    \cline{2-5}
                                    & \multirow{4}{*}{Yahoo!}   & \multirow{2}{*}{CCV}          & FC + CONV     & 0.484 \\
                                    &                           &                               & FC            & 0.497 \\
                                                                \cline{3-5}
                                    &                           & \multirow{2}{*}{CCV + Yahoo!} & FC + CONV     & 0.494 \\
                                    &                           &                               & FC            & 0.499 \\
                                    \cline{2-5}
                                    & \multirow{4}{*}{ILSVRC}   & \multirow{2}{*}{CCV}          & FC + CONV     & 0.489 \\
                                    &                           &                               & FC            & 0.490 \\
                                                                \cline{3-5}
                                    &                           & \multirow{2}{*}{CCV + ILSVRC} & FC + CONV     & 0.499 \\
                                    &                           &                               & FC            & \textbf{0.508} \\
      \hline
      \multirow{11}{*}{5-Conv.}     & \multirow{3}{*}{Random}   & CCV                           & FC + CONV     & 0.469 \\
                                                                \cline{3-5}
                                    &                           & CCV + Yahoo!                  & FC + CONV     & 0.482 \\
                                                                \cline{3-5}
                                    &                           & CCV + ILSVRC                  & FC + CON      & 0.486 \\
                                    \cline{2-5}
                                    & \multirow{4}{*}{Yahoo!}   & \multirow{2}{*}{CCV}          & FC + CONV     & 0.483 \\
                                    &                           &                               & FC            & 0.510 \\
                                                                \cline{3-5}
                                    &                           & \multirow{2}{*}{CCV + Yahoo!} & FC + CONV     & 0.513 \\
                                    &                           &                               & FC            & 0.506 \\
                                    \cline{2-5}
                                    & \multirow{4}{*}{ILSVRC}   & \multirow{2}{*}{CCV}          & FC + CONV     & 0.518 \\
                                    &                           &                               & FC            & 0.561 \\
                                                                \cline{3-5}
                                    &                           & \multirow{2}{*}{CCV + ILSVRC} & FC + CONV     & 0.559 \\
                                    &                           &                               & FC            & \textbf{0.565} \\
      \hline
      \end{tabu}
      \caption{CCV}
  \end{subtable}
  \begin{subtable}{0.6\textwidth}
        \begin{tabu}{c|c|c|c|c}
        \hline
        Depth                       & Initialization            & Training Set                      & Update Policy & MAP \\
        \hline
        \multirow{11}{*}{2-Conv.}   & \multirow{3}{*}{Random}   & TRECVID                           & FC + CONV     & 0.620 \\
                                                                \cline{3-5}
                                    &                           & TRECVID + Yahoo!                  & FC + CONV     & 0.628 \\
                                                                \cline{3-5}
                                    &                           & TRECVID + ILSVRC                  & FC + CONV     & 0.671 \\
                                    \cline{2-5}
                                    & \multirow{4}{*}{Yahoo!}   & \multirow{2}{*}{TRECVID}          & FC + CONV     & 0.659 \\
                                    &                           &                                   & FC            & 0.700 \\
                                                                \cline{3-5}
                                    &                           & \multirow{2}{*}{TRECVID + Yahoo!} & FC + CONV     & 0.692 \\
                                    &                           &                                   & FC            & 0.708 \\
                                    \cline{2-5}
                                    & \multirow{4}{*}{ILSVRC}   & \multirow{2}{*}{TRECVID}          & FC + CONV     & 0.688 \\
                                    &                           &                                   & FC            & 0.717 \\
                                                                \cline{3-5}
                                    &                           & \multirow{2}{*}{TRECVID + ILSVRC} & FC + CONV     & \textbf{0.751} \\
                                    &                           &                                   & FC            & 0.746 \\
      \hline
      \multirow{11}{*}{5-Conv.}     & \multirow{3}{*}{Random}   & TRECVID                           & FC + CONV     & 0.645 \\
                                                                \cline{3-5}
                                    &                           & TRECVID + Yahoo!                  & FC + CONV     & 0.674 \\
                                                                \cline{3-5}
                                    &                           & TRECVID + ILSVRC                  & FC + CON      & 0.737 \\
                                    \cline{2-5}
                                    & \multirow{4}{*}{Yahoo!}   & \multirow{2}{*}{TRECVID}          & FC + CONV     & 0.718 \\
                                    &                           &                                   & FC            & 0.749 \\
                                                                \cline{3-5}
                                    &                           & \multirow{2}{*}{TRECVID + Yahoo!} & FC + CONV     & 0.767 \\
                                    &                           &                                   & FC            & 0.745 \\
                                    \cline{2-5}
                                    & \multirow{4}{*}{ILSVRC}   & \multirow{2}{*}{TRECVID}          & FC + CONV     & 0.767 \\
                                    &                           &                                   & FC            & 0.843 \\
                                                                \cline{3-5}
                                    &                           & \multirow{2}{*}{TRECVID + ILSVRC} & FC + CONV     & 0.833 \\
                                    &                           &                                   & FC            & \textbf{0.859} \\
      \hline
      \end{tabu}
      \caption{TRECVID MED11}
  \end{subtable}

  \label{tab:ccv_all}
\end{table*}

In this section, we examine the approach of augmenting video datasets with static image datasets.
In practice, we subsample roughly the same number of images as video frames (0.4 million) from the image datasets
and mix these images with the video datasets to create a new training set.
The results are shown in Table~\ref{tab:ccv_all}.
Although the two datasets are from very different sources (static-image v.s. video),
augmenting the video datasets indeed improves the performance.
This may be explained by the first convolution layer kernels, as in Fig.~\ref{fig:videokernels},
which shows that by augmenting video dataset, 
the network successfully learns the high frequency signals that are ignored when training the network on \textsl{CCV} only.
These signal are known to be important, and ignoring them may degrade the generalizability.

\subsection{Transfer Mid-level Features}

\begin{figure}[t]
\centering
\includegraphics[width=\linewidth]{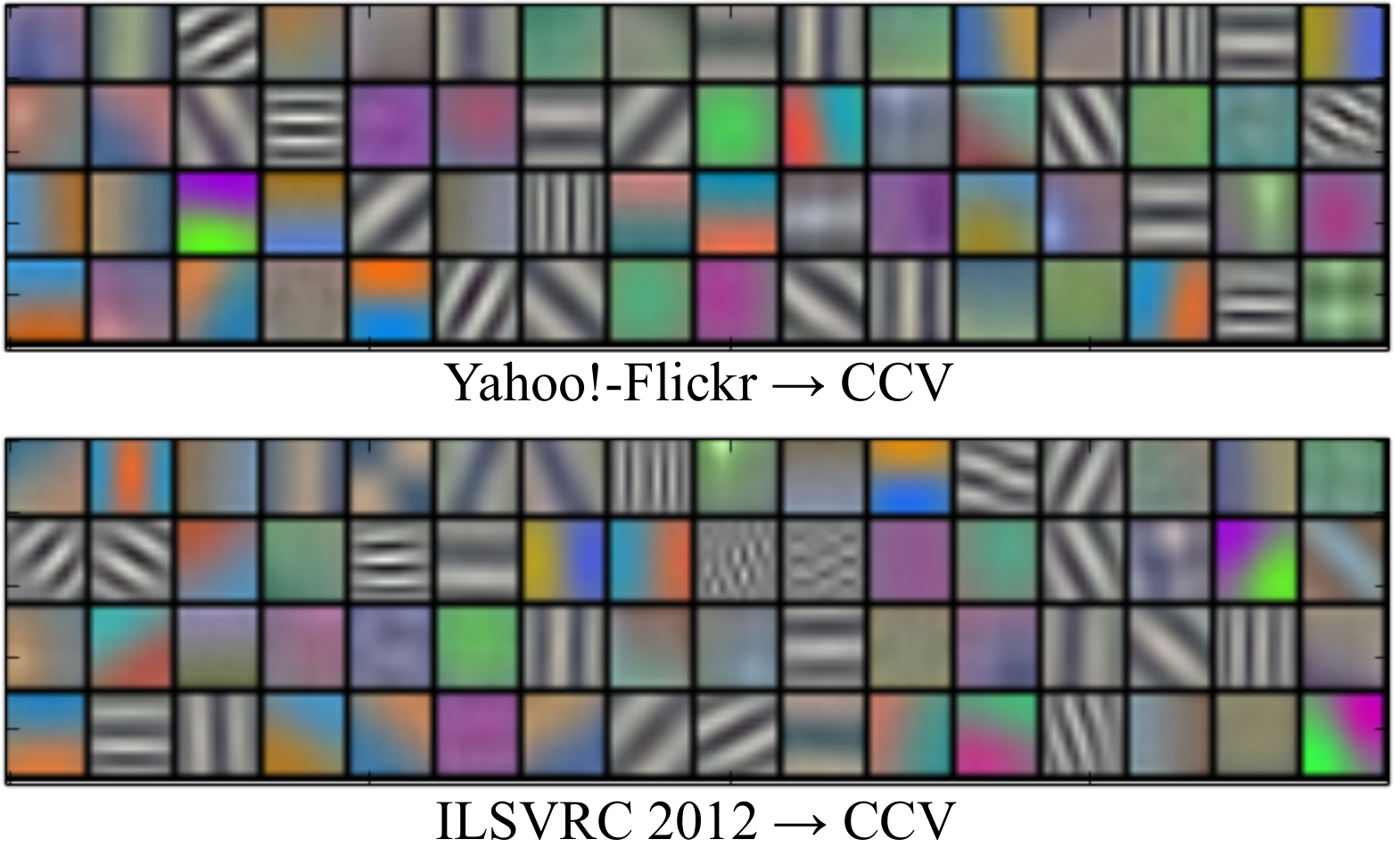}
\caption{\label{fig:transferkernels} The first convolution layer kernels of networks using a pre-trained network as initialization.
Although the kernels are both learned from the \textsl{CCV} dataset,
they show different visual patterns which are more similar to their initialization points respectively.
Note the clear 1-to-1 correspondence between the kernels above and those in Fig.~\ref{fig:imagekernels}.
In fact, most of the kernels do not change significantly after fine-tuning with \textsl{CCV} datasets,
which indicates the patterns learned from either \textsl{Yahoo!-Flickr} or \textsl{ILSVRC2012} are helpful for recognizing \textsl{CCV} videos, i.e. datasets of different domains.
}
\end{figure}

In this section, we examine the approach of transferring mid-level features.
In practice, we initialize the network for video recognition by the network trained on \textsl{Yahoo!-Flickr} and \textsl{ILSVRC2012} respectively.
During training, we either update all the parameters in the network or update only the fully connected layers 
and keep the convolution kernels unchanged.
The reason why we keep the convolution kernels unchanged is to reduce the number of learnable parameters to avoid overfitting as motivated by the convolution regularization.
If the lower layer convolution kernels do learn important patterns such as lines or corners,
they should be similar and reusable over different datasets, so keeping them unchanged should not degrade the performance.
The result are in Table~\ref{tab:ccv_all}.
Initializing the network with pre-trained networks improves the performance significantly,
and updating only the fully connected layers is better than updating all parameters.
This indicates that when the training data is not enough, 
avoiding to update the low level features in DCNs will reduce the overfitting problem and lead to better performance.
These results are identical with that in \cite{Karpathy:2014cvpr}, 
which shows that update only the fully connected layers lead to better recognition accuracy,
although we use totally different data and different network structures.
It is also consistent with \cite{Girshick:2014cvpr},
which shows that fully connected layers in DCNs are less generalizable than convolution layers
and therefore require fine-tuning.
Note that the networks learn different sets of convolution kernels using different initialization,
as shown in Fig.~\ref{fig:transferkernels}, while both of them achieve good performance.

Finally, we combine the two transfer learning approaches.
That is, we initialize the network with the one trained on image datasets
and fine-tune the network with both images and video frames.
The recognition performance is further improved by the combination.
It is interesting to see the difference between different image datasets used to bootstrap the video recognizer.
While the two image datasets lead to similar performance when 2-convolution-layer network is used,
the \textsl{ILSVRC2012} benefit more from the additional depths,
because the more precise label information provides better supervision and enables learning a more complex model.
This indicates that the network benefits from any additional supervision, 
even if it is independent of the target problem.
Nevertheless, the weakly supervised \textsl{Yahoo!-Flickr} also boosts the video recognition performance
and does not require the expensive human annotations for ground truth.
The only annotated data in the entire learning process is the video dataset, 
which requires much less human intervention and annotation overhead.

\section{Conclusions}
\label{sec:conclusion}
In this work, we apply DCNs on unconstrained video recognition, where the networks are used as frame-based recognizers.
Our preliminary study shows that one of the most significant obstacles for training a DCN is 
the requirement for a large amount of training samples.
Without enough training samples, the networks are highly prone to overfitting which leads to poor performance.
This problem is especially important for video recognition,
because videos with ground truth are scarce and hard to obtain.

To overcome the problem, we train DCNs with transfer learning from images to videos.
The image corpus can be weakly labeled, which is widely available in various social media such as Flickr and Instagram.
These rich image samples can help to learn more robust recognizers on the video frames, even though they are from different domains.
The transfer learning process makes training DCN with scarce training data possible,
where we achieve reasonable performance using only 4k videos for training.
Because weakly labeled datasets are good enough for supervised pre-training without harming the performance,
we are exempt from the requirement of collecting and annotating a large dataset as in previous researches in DCN.

Our preliminary study also reveals the correlation between meta-parameters and performance given different dataset properties.
In particular, we study the effect of depth and image resolution, because these factors have significant impact on the computation cost.
The results indicate that high resolution images always yield better performance,
but it is more important for object level recognition compared with scene level recognition.
The results also show that additional depths in the network may not always be helpful for performance;
in fact, deep networks may sometimes perform worse than shallow ones.
These studies not only help us to select the meta-parameters used for video recognition,
but also provide hints for future researcher and facilitate the research in DCN.

In our future work, we would like to investigate the possibility of unsupervised pre-training for DCN.
Our current approach still requires a large labeled corpus, and we would like to further eliminate the requirement.

\ifCLASSOPTIONcaptionsoff
  \newpage
\fi



%
\bibliographystyle{IEEEtran}
\bibliography{tcsvt2014su}




%






\end{document}